\documentclass{article}

\usepackage{microtype}
\usepackage{graphicx}
\usepackage{subfigure}
\usepackage{booktabs} 

\usepackage{hyperref}

\usepackage[utf8]{inputenc} 
\DeclareUnicodeCharacter{2212}{-}
\usepackage[T1]{fontenc}    
\usepackage{hyperref}       
\usepackage{url}            
\usepackage{booktabs}       
\usepackage{amsfonts}       
\usepackage{nicefrac}       
\usepackage{microtype}      
\usepackage{pgfplots}
\usepackage{xcolor}
\usepackage{amsmath}
\usepackage{amsthm}
\usepackage{amsfonts}
\usepackage{amsmath,amsfonts,amsthm,bm} 
\usepackage{thmtools}
\usepackage{thm-restate}
\usepackage{hyperref}
\usepackage{mathtools}
\usepackage{url}
\usepackage{graphicx}
\graphicspath{ {./images/} }
\usepackage{pgf}
\usepackage{tikz}
\usetikzlibrary{arrows,automata,positioning}
\usepackage{wrapfig}

\DeclarePairedDelimiter\br{(}{)}
\DeclarePairedDelimiter\brs{[}{]}
\DeclarePairedDelimiter\brc{\{}{\}}
\DeclarePairedDelimiter\abs{\lvert}{\rvert}
\DeclarePairedDelimiter\norm{\lVert}{\rVert}

\newcommand{\ntext}[1]{{\text{\normalfont#1}}}
\newcommand{\normal}[1]{\mathcal{N}\br*{#1}}
\newcommand{\E}{\mathbb{E}} 
\newcommand{\R}{\mathbb{R}} 

\newcommand{\gammaEval}{\gamma_{e}}
\newcommand{\Vhat}{\hat{V}}
\newcommand{\Qhat}{\hat{Q}}

\newcommand{\VhatT}{\Vhat_\theta}
\newcommand{\QhatT}{\Qhat_\theta}
\newcommand{\VhatTi}{\Vhat_{\theta_i}}
\newcommand{\QhatTi}{\Qhat_{\theta_i}}
\newcommand{\QhatPi}{\Qhat_{\phi_i}}

\newcommand{\VhatPi}{\Vhat_{\phi_i}}
\newcommand{\VhatBackup}{\Vhat_{\Tilde{\theta}}}
\newcommand{\GradVhatT}{\nabla \hat{V}_\theta}
\newcommand{\GradVhatTi}{\nabla \hat{V}_{\theta_i}}
\newcommand{\GradQhatTi}{\nabla \hat{Q}_{\theta_i}}
\newcommand{\GradQhatPi}{\nabla \hat{Q}_{\phi_i}}

\newcommand{\GradVhatPi}{\nabla \hat{V}_{\phi_i}}
\newcommand{\Rmax}{R_{\ntext{max}}}

\newcommand{\States}{\mathcal{S}}
\newcommand{\Actions}{\mathcal{A}}
\newcommand{\TransProb}{P}
\newcommand{\RewardProb}{R}
\newcommand{\Distrib}{\mathcal{M}}
\newcommand{\MDP}{\mathcal{M}}
\newcommand{\Niter}{ N_\ntext{iter}}

\newcommand{\RegTerm}{\Psi}

\newtheorem{proposition}{Proposition}

\usepackage[accepted]{icml2020}

\icmltitlerunning{Discount Factor as a Regularizer in Reinforcement Learning}

\begin{document}

\twocolumn[
\icmltitle{Discount Factor as a Regularizer \\in Reinforcement Learning}



\icmlsetsymbol{equal}{*}

\begin{icmlauthorlist}
\icmlauthor{Ron Amit}{Technion}
\icmlauthor{Ron Meir}{Technion}
\icmlauthor{Kamil Ciosek}{Microsoft}
\end{icmlauthorlist}

\icmlaffiliation{Technion}{The Viterbi Faculty of Electrical Engineering, Technion - Israel Institute of Technology, Haifa, Israel}
\icmlaffiliation{Microsoft}{Microsoft Research, Cambridge, UK}

\icmlcorrespondingauthor{Ron Amit}{ronamit@campus.technion.ac.il}
\icmlcorrespondingauthor{Ron Meir}{rmeir@ee.technion.ac.il}
\icmlcorrespondingauthor{Kamil Ciosek}{Kamil.Ciosek@microsoft.com}

\icmlkeywords{Machine Learning, ICML}

\vskip 0.3in
]



\printAffiliationsAndNotice{}  

\begin{abstract}
Specifying a Reinforcement Learning (RL) task involves choosing a suitable planning horizon, which is typically modeled by a discount factor. It is known that applying RL algorithms with a lower discount factor can act as a regularizer, improving performance in the limited data regime. Yet the exact nature of this regularizer has not been investigated. In this work, we fill in this gap. For several Temporal-Difference (TD) learning methods, we show an explicit equivalence between using a reduced discount factor and adding an explicit regularization term to the algorithm's loss.
Motivated by the equivalence, we empirically study this technique compared to standard $L_2$ regularization by extensive experiments in discrete and continuous domains, using tabular and functional representations.
Our experiments suggest the regularization effectiveness is strongly related to properties of the available data, such as size, distribution, and mixing rate.

\end{abstract}

\section{Introduction}

The ability to perform well in new and unfamiliar situations following a limited learning experience is a hallmark of human intelligence. Similarly, the generalization ability of Reinforcement Learning (RL) algorithms is often measured by expected performance achieved by the agent in a Markov Decision Process (MDP) after being exposed to a limited amount of training data. Developing RL agents that generalize well is a longstanding challenge \citep{boyan1995generalization, sutton1996generalization} that has recently been gaining more attention  \citep{cobbe2018quantifying, zhang2018study, zhang2018dissection, wang2019generalization, zhao2019investigating}. In particular, generalization is critical for successfully deploying RL agents that were trained in a simulator in complex real-world scenarios that contain elements not seen in the simulation. 

There are several known approaches for improving generalization in RL. 
 Selecting an appropriate function approximation model is one way to facilitate generalization across states and actions \cite{boyan1995generalization}. Regularization methods can further improve the generalization capacity. 
For example, it is very common to perform regularization in policy space by encouraging policies with high entropy \citep{williams1992simple, mnih2016asynchronous, ahmed2019understanding, vieillard2020leverage}. Our focus is instead on policy evaluation. Traditionally, there have been two common approaches to such regularization.
First, one can use traditional regularization methods from supervised learning to estimate the value function. Most commonly, this means adding an $L_2$ or $L_1$ penalty on the parameters of the value function (critic) \citep{kolter2009regularization,liu2012regularized, dann2014policy, lillicrap2015continuous, cobbe2018quantifying, liu2019regularization}.
Second, one can apply indirect regularization by running the learning algorithm with a discount factor lower than specified by the task. 
We refer to this method as \textit{discount regularization.}
By focusing learning on short-term gains, this approach may improve generalization by reducing variance \citep{petrik2009biasing, jiang2015dependence, jiang2015abstraction, franccois2019overfitting, vanseijen2019using}. This leads to the question:
\vspace*{-0.5em}
\begin{center}{ \emph{ What are the factors that influence the effectiveness of discount regularization?} }
\end{center}

This paper contributes to answering this question in three ways. First, for a few variants of TD learning, we show an equivalence between using a reduced discount and \textit{activation regularization}, a technique used to train Recurrent Neural Networks (RNNs) \citep{merity2017revisiting, merity2018regularizing, herold2018improving}. 

Second, we empirically investigate the effectiveness of discount regularization  in both tabular MDPs and large scale continuous control benchmarks. We show the benefit of discount regularization is strongly linked to the number of samples, uniformity of the state visitation and mixing rate of the data collection. Generally, discount regularization is more effective when data is limited, data distribution is highly uniform, and the mixing rate is low.
In general, we fond discount regularization and $L_2$ regularization have similar performance in tabular settings,  but vary in some function approximation settings.



Section \ref{sect:Background} provides background on TD learning.
In Section  \ref{sect:equiv} we formalize the equivalence between using an artificially lowered discount and activation regularization.  
In Sections \ref{sect:tabular_emprical_demonstration} and \ref{sect:deep_experiments}  we investigate our predictions empirically in tabular and deep RL benchmarks respectively.
Section \ref{sect:Related} discusses related work.
    
\section{Background} \label{sect:Background}
\subsection{Problem Setting}
An MDP \citep{bellman1957markovian} is defined as a tuple $\MDP := (\States, \Actions, \TransProb, \RewardProb, \mu)$, where $\States$ is the state set, $\Actions$ is the action set, $\TransProb : \States \times \Actions \rightarrow  \Distrib(\States)$ is the transition probability function,  $\Distrib(\States)$ is the set of distributions over $\States$,  $\RewardProb: \States \times \Actions \rightarrow  \Distrib([0,\Rmax])$ is the reward distribution function, $\Distrib([0,\Rmax])$ is the set of distributions supported on $[0,\Rmax]$ and $\mu \in \Distrib(\States)$ is the initial state distribution.
A Markovian stationary policy is defined by a mapping $\pi: \States \rightarrow \Distrib(\Actions)$.
At each time-step $t$ the agent draws an action $a_t$ from $\pi(s_t)$ where $s_t$ is the current state. The agent then receives a random reward $r_t \sim \RewardProb(s_t,a_t)$ and transitions to the next state $s_{t+1}$ drawn from $P(s_t,a_t)$. 
This process produces a (possibly infinite) trajectory $\tau := (s_0,a_0,r_0,s_1,a_1,r_1,...)$.
Given a discount factor $\gamma \in [0,1]$ the value function at state $s$ is defined by the expected discounted return $V_\gamma^\pi(s):=\E_{\tau:s_0=s} \brs*{\sum_{t=0}^{\infty} \gamma^t r_t | s_0=s}$. 
Similarly we define the Q-function given state $s$ and action $a$ as $Q_\gamma^\pi(s,a):=\E_{\tau:s_0=s,a_0=a} \brs*{\sum_{t=0}^{\infty} \gamma^t r_t | s_0=s,a_0=a}$. 

In our setting, the agent is allowed to observe a limited number of samples of trajectories generated from $\MDP$. We define a sample as a single transition $(s,a,r,s')$, where $s$ is the current state, $a$ is the action taken, $r$ is the immediate reward , and $s'$ is the next state.
We investigate two types of goals:  \textit{policy evaluation} and \textit{control}.
In policy evaluation the agent is given a fixed policy $\pi$ and aims to estimate $V_{\gammaEval}^\pi(s)$ where $\gammaEval \in [0,1]$ is the \textit{evaluation discount factor}.
In the control setting the agent aims to find a policy  $\pi$ that maximizes the expected return $\E_{\tau:\pi} \brs*{\sum_{t=0}^{\infty} \gammaEval^t r_t}$. In this paper we investigate control algorithms that include policy evaluation as one constituent component.

We consider policy evaluation with function approximation, where the estimated value function is chosen from a parametric family $\brc*{\VhatT : \States \rightarrow \R | \theta \in \R^d}$. We assume the functions in this family are differentiable  w.r.t.~$\theta$. The tabular setting can be considered as a special case with $\VhatT(s):=\theta_s, \theta \in \R^{\abs*{\States}}$.
\subsection{Temporal-Difference Learning}
The Temporal-Difference (TD) learning algorithm family \citep{sutton1988learning} is used for efficient policy evaluation.
While our insights apply to a wide range of TD methods, we focus our discussion on TD(0) as a representative algorithm. We address the $m$-step variant and the SARSA algorithm in Appendices \ref{sect:Sarsa_Equiv} and \ref{sect:EquivMstep}.
We will consider a batch setting, in which the task is to estimate the value function  $V_{\gammaEval}^\pi(s)$ of a known policy $\pi$ given samples from trajectories generated by interaction of $\pi$ with the MDP $\MDP$.
We assume the finite data setting, i.e, we are given a data set $D$  of $n$ samples. Since we are interested in effects of finite sample size and not a finite number of iterations, we choose to focus on the  batch rather than an online setting. In the batch setting, we can reuse each sample in the data set for many iterations.

Algorithm \ref{alg:TD0} is a generic form of a regularized batch TD(0) algorithm.
In the special case of the standard non-regularized TD(0), there is no added regularization term ($\RegTerm \equiv 0$), there is no reward scaling $\xi = 1$, and the discount factor used is the one desired in the problem definition ($\gamma=\gammaEval$).
The algorithm is initialized at some initial parameters $\theta_0$ and takes steps aiming to minimize $\E_{(s,a,r,s') \sim D} \brc*{\br*{r + \gamma \VhatBackup(s') - \VhatT(s)}^2 + \RegTerm}$, where the expectation is w.r.t.~a the empirical distribution over  samples $D$.
Similarly to Stochastic Gradient Descent (SGD), in each iteration only one transition $(s,a,r,s')$ is sampled from $D$ to approximate the full gradient.
For stability considerations, instead of the standard gradient, the algorithm computes a \textit{`semi-gradient'} \citep{sutton2018reinforcement}, i.e. the next state value estimate, $\VhatBackup(s')$, is fixed.
The learning rate $\alpha_i \in \R^+$ is usually set to be monotonically decaying at rate $O(1/i)$ in table-lookup settings and scaled automatically \citep{kingma-adam} in deep learning settings.

\begin{algorithm} 
  \caption{Generic Regularized Batch TD(0)}
  \label{alg:TD0}
\begin{algorithmic}
  \STATE {\bfseries Hyper-parameters:} $\gamma \in [0, \gammaEval]$,  
  $\xi \in \R^+$  (global reward scaling), $\RegTerm$ (regularization function) 
  \STATE {\bfseries Input:}   $D$
    \FOR{ $i=0,1,..., \Niter - 1$}
       \STATE  Get uniformly random $(s,a,r,s')$ from $D$ 
       \STATE  $\theta_{i+1} :=  \theta_{i} + \alpha_i \br*{\xi r + \gamma \VhatTi(s') - \VhatTi(s)} \GradVhatTi(s) - \alpha_i  \nabla \br{\RegTerm}.$  
    \ENDFOR
\end{algorithmic} 
\end{algorithm}

We are interested in the result in the limit of an infinite number of iterations $\Niter \rightarrow \infty$ for which all samples from $D$ are used infinitely often.
Note that since we are dealing with finite data, convergence to the true value is not guaranteed even for  $\gamma:=\gammaEval$.
We refer to the discount factor $\gamma \in (0,1)$ used by the algorithm as the \textit{guidance discount factor} \citep{jiang2015dependence}.
In this paper we study the regularizing effect of using $\gamma$ lower than the evaluation discount factor $\gammaEval$  and compare it with other regularization methods.

\paragraph{Q-function evaluation.}
In many cases (e.g., control)  we are interested in estimating the action-value $Q^\pi$ function rather than $V^\pi$.
The naive variant of TD(0) for estimating the Q-function is the SARSA(0)   algorithm \citep{rummery1994line}. In Appendix \ref{sect:Sarsa_Equiv}, we also discuss a variant called Expected SARSA(0) \citep{sutton1998introduction} which utilizes knowledge of $\pi$ to perform lower variance updates \citep{van2009theoretical}.

\paragraph{Policy iteration.} In our work, we investigate control algorithms that fit the policy iteration framework, i.e, algorithms that alternate between policy evaluation and policy improvement. Specifically, we investigate algorithms that use TD-style policy evaluation.
Many control RL algorithms fit this framework, including modern actor-critic methods such as DPG \citep{silver2014deterministic}, SAC \citep{haarnoja2018soft}, DDPG \citep{lillicrap2015continuous}, and Twin Delayed DDPG (TD3) \citep{fujimoto2018addressing}, which is investigated in the experiments section.

\section{Discount Regularization in TD Learning} \label{sect:discount_reg}

\subsection{Equivalence of Reduced Discount Factor and Activation Regularization} \label{sect:equiv}


In this section, we formulate the equivalence between TD learning with a reduced discount and TD learning with a high discount with an added regularization term.  The equivalence will provide insights about the effectiveness of discount regularization in various settings.

For simplicity of presentation, we first show that TD(0) with  guidance discount factor $\gamma < \gammaEval$  is equivalent to an added activation regularization term to the standard $\gammaEval$-discounted update.
Analogous results can obtained for SARSA (Appendix \ref{sect:Sarsa_Equiv}), $m$-step TD (Appendix \ref{sect:EquivMstep}) and LSTD (Appendix \ref{sect:EquivLSTD}).  
The proof is in Appendix \ref{sect:proof_td0}.


\begin{proposition} \label{prop:Equiv_TD0}
Let $\theta_1, \theta_2, \dots$ be  the parameters produced by Algorithm \ref{alg:TD0} using a discount factor $\gamma < \gammaEval$, with $\xi=1, \RegTerm \equiv 0$, initial parameters $\theta_0$ and learning rate $\alpha_i$. 
The algorithm, produces the same sequence of parameters  $\theta_1, \theta_2 \dots$ if it is run with the discount factor $\gamma =\gammaEval$, but with added regularization function $\RegTerm(s, \theta) :=  \lambda \br*{\VhatT(s)}^2$, $\lambda := \frac{\gammaEval - \gamma}{2\gamma}$ , 
reward scaling $\xi:= \frac{\gammaEval}{\gamma}$,
learning rate  $\alpha_i' :=  \frac{\gamma}{\gammaEval} \alpha_i$
and the same initial parameters $\theta_0$.
\end{proposition}


Proposition \ref{prop:Equiv_TD0} implies that running TD(0) with a reduced discount factor is equivalent to minimizing the objective
 $\E_{(s,a,r,s') \sim D} \brc*{\br*{ \xi r +  \gammaEval \VhatBackup(s') - \VhatT(s)}^2 + \lambda  \br*{ \VhatT(s)}^2}$.
We refer to the added regularization term as \textit{activation regularization}\footnote{This naming relates to activation regularization in RNNs, which refers to $L_2$ penalty on the RNN activations, rather than on the weights of the network in standard $L_2$ regularization \citep{merity2017revisiting, merity2018regularizing}}.
In TD(0), this term is the mean value of the square of the learned value function over the distribution of observed states  $\lambda  \E_s   \br*{\VhatT(s)}^2$. In the SARSA algorithm we have a similar term  $\lambda  \E_{(s,a)} \br*{\QhatT(s,a)}^2$ (see Appendix \ref{sect:Sarsa_Equiv}).
This term penalizes large value estimates and therefore encourages consistent value estimates across state-action pairs, which may encourage generalization by reducing the effect of spurious approximation errors.
Reducing  $\gamma$ increases the factor of the equivalent regularization term  $\lambda := \frac{\gammaEval - \gamma}{2\gamma}$.



We can get a more explicit form for the activation regularization  when using a tabular function or a linear approximation with orthogonal features, where the activation regularization term is equal to a weighted $L_2$ norm  on the parameters.
Define $\VhatT(s) := \phi(s)^\top \theta$ for some fixed feature mapping $\phi$ and some weight vector $\theta \in \R^k$.
Assume orthogonal features, i.e, that we have\footnote{The expectation is w.r.t a uniform distribution over the samples $(s,a,r,s') \in D$.} $\E_s\left [  \phi(s) \phi(s)^\top \right] = \Lambda$ for some diagonal matrix $\Lambda$. This assumption holds for the tabular case for which $\phi(s) = e_s$ where $e_s$ is the standard basis of $\R^{|\States|}$.
The activation regularization term can be written as
\begin{align} \label{eq:act_reg_linear_case}
\lambda \E_s \br*{\VhatT(s)}^2 &= \lambda  \E_s\left [   ( \theta^\top \phi(s) \phi(s)^\top \theta    ) \right] \\
&= \lambda \theta^\top \Lambda \theta    = \lambda  \| \theta \| _\Lambda^2 . \nonumber
\end{align}

If, in addition, the features are also \textit{orthonormal}, i.e, $ \Lambda =  E_s\left [ \phi(s) \phi(s)^\top \right] =  \mathbb{I}_{k \times k}$ then  the activation regularization term becomes equivalent to the an $L_2$ regularization term $\norm{\theta}_2^2.$
For example, this case applies for tabular representation when the data distribution is uniform across states.
Note that even in this case, if we want discount regularization to be equivalent to   $L_2$  regularized algorithm with  $\gamma:=\gammaEval$, Proposition \ref{prop:Equiv_TD0} claims that we should adjust the reward scaling and learning rate: $\xi := \frac{\gammaEval}{\gamma}, \alpha'_i:=  \frac{\gamma}{\gammaEval} \alpha_i$  (i.e, the inverse transformation to the one described in the proposition).

Proposition \ref{prop:Equiv_TD0} showed that a reduced discount is equivalent to adding a activation regularization term $\lambda \E_{s \sim D} \br*{\VhatT(s)}^2 $ to the learning objective.
Notice that this term is sensitive to the distribution over the observed states.
For example, in  the tabular case, $\VhatT(s) := \theta_s$, 
the activation regularization term is simplified to $\lambda \E_{s \sim D}  \theta_s^2$.
This form demonstrates that states that are visited less often are less regularized, i.e, the regularization factors for these states are lower. If a state is not visited at all, the value estimation for this state is not regularized at all.  

This phenomenon raises a concern that activation regularization (or equivalently small discount) may be less helpful for generalization state visitation is farther from a uniform distribution.
In Section \ref{sect:tabular_emprical_demonstration} we will demonstrate empirically that discount regularization is indeed less beneficial when the data distribution is highly non-uniform.

\section{Empirical Demonstrations} \label{sect:emprical_demonstration}

The goal of the of the experiments in this section is to investigate the following questions\footnote{Code for all the experiments is available at:  \url{https://github.com/ron-amit/Discount_as_Regularizer}.}.
Can reducing the discount factor improve generalization performance with TD learning? 
How is the optimal discount factor related to data size?
What is the effect of data uniformity and mixing rate? 
What is the benefit of discount regularization compared to $L_2$ regularization (in both tabular and function approximation settings)?

\subsection{Tabular Experiments} \label{sect:tabular_emprical_demonstration}

\newcommand{\tabularPlotScale}{0.26}
\begin{figure*}[t]
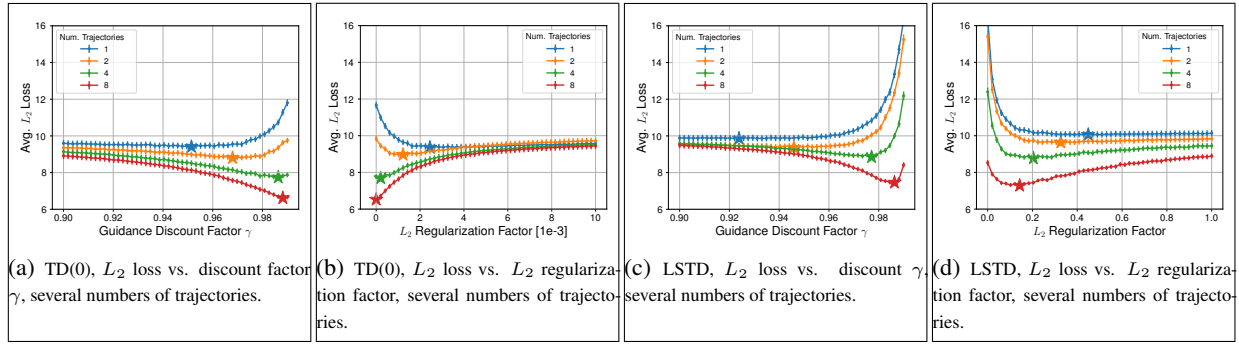

\centering
 \frame{
\subfigure[\scriptsize  {TD(0), $L_2$ loss vs. discount factor $\gamma$, several numbers of trajectories.\newline}]{
  \scalebox{\tabularPlotScale}{\input{images/2020_06_15_13_52_52_PolEval_TD_L2Loss_DiscountReg.pgf}}
  \label{fig:PolEval_TD_L2Loss_DiscountReg}
}}
 \frame{
\subfigure[\scriptsize  {TD(0),  $L_2$ loss vs. $L_2$ regularization factor, several numbers of trajectories.}]{
  \scalebox{\tabularPlotScale}{\input{images/2020_06_15_14_52_20_PolEval_TD_L2Loss_L2Reg.pgf}}
  \label{fig:PolEval_TD_L2Loss_L2Reg}
}}
 \frame{
\subfigure[\scriptsize  {LSTD, $L_2$ loss vs. discount $\gamma$, several numbers of trajectories.\newline}]{
  \scalebox{\tabularPlotScale}{\input{images/2020_06_15_17_04_47_PolEval_LSTD_L2Loss_DiscountReg.pgf}}
  \label{fig:PolEval_LSTD_L2Loss_DiscountReg}
}}
 \frame{
\subfigure[\scriptsize  {LSTD, $L_2$ loss vs. $L_2$ regularization factor, several numbers of trajectories.}]{
  \scalebox{\tabularPlotScale}{\input{images/2020_06_16_00_52_40_PolEval_LSTD_L2Loss_L2Reg.pgf}}
  \label{fig:PolEval_LSTD_L2Loss_L2Reg}
}}
\caption{\textbf{Tabular experiments - effect of dataset size.} Loss vs. regularization factor for different regularizers, and algorithms, averaged over $1000$ MDP instances.
In each figure, the curves correspond to different number of samples per episode. The star shapes mark the minimum of the curve. Error bars represent $95\%$ confidence interval.}
\label{fig:Tabular_L2_Loss}
\end{figure*}

We first investigate the effectiveness and discount regularization in various setting we conducted a simple GridWorld experiment.

In the GirdWorld environment the state space is a $4 \times 4$ grid, and the agent can move to along the gird. In each experiment, we randomly choose a `goal state' to be assigned with a high reward mean. The other reward means and the transition probabilities are also generated randomly.
The full details of the experiment appear in Appendix \ref{sect:additional_details_tabular}.

We first experiment in a batch policy evaluation setting, for a fixed uniform policy.
The evaluation metric we use  is the $L_2$ distance of the estimated value from the true value $V^{\pi}_{\gammaEval}$,  $\norm{\Vhat - V_{\gammaEval}^{\pi}}_2$, where $ V_{\gammaEval}^{\pi}$ is evaluated with  $\gammaEval=0.99$. 
In our first set of experiments we generate the data by simulating trajectories starting at a random initial state and following a uniform policy for $50$ time-steps. We varied the number of trajectories to change the sample size. 

The results are summarized in Figure \ref{fig:Tabular_L2_Loss}. Each plot shows the average loss across $1000$ MDP instances and the $95\%$ confidence intervals. 
In Figure \ref{fig:PolEval_TD_L2Loss_DiscountReg}, we clearly see that using a smaller discount factor $\gamma < \gammaEval$ can significantly improve performance when the available data set is small. This corresponds to our observation that a smaller discount is equivalent to a stronger activation regularization term.
In Figure \ref{fig:PolEval_TD_L2Loss_L2Reg}, we see the effect of $L_2$ regularization with no discount regularization ($\gamma = \gammaEval$). The results show $L_2$ regularization achieves similar performance gain as  discount regularization.
Figures \ref{fig:PolEval_LSTD_L2Loss_DiscountReg} and \ref{fig:PolEval_LSTD_L2Loss_L2Reg} show the corresponding results with the LSTD algorithm.
In contrast to TD(0), for LSTD we see that regularization can improve performance for all data set sizes, and the loss when not using regularization is higher.

In some case, the actual values of the estimates are less important than the relative rankings  of the values of states. 
Therefore, we repeated the experiment with a loss function that compares state rankings (see Appendix \ref{sect:ranking_loss}).  The results show similar behaviour as with the $L_2$ loss.

We have seen that regularization is more helpful when the data size is limited.
But there are other properties of the data that indicate that regularization may be more effective. 
Next, we will investigate the influence of the uniformity of the data distribution and of the mixing rate of the data generating process.

\paragraph{Influence of the uniformity of the data state distribution.}
We consider a batch setting, where the state-action tuples are drawn independently from fixed distributions (while the reward, next state, and next action are drawn according to the environment stochasticity and the evaluated policy).  
 To measure the uniformity of the distributions, we evaluated the total variation distance from a uniform distribution.
 In each experiment repetition, we randomly generated distributions with various distances via rejection sampling. The data consists of $400$ sampled tuples.
 
\begin{figure*}[t]
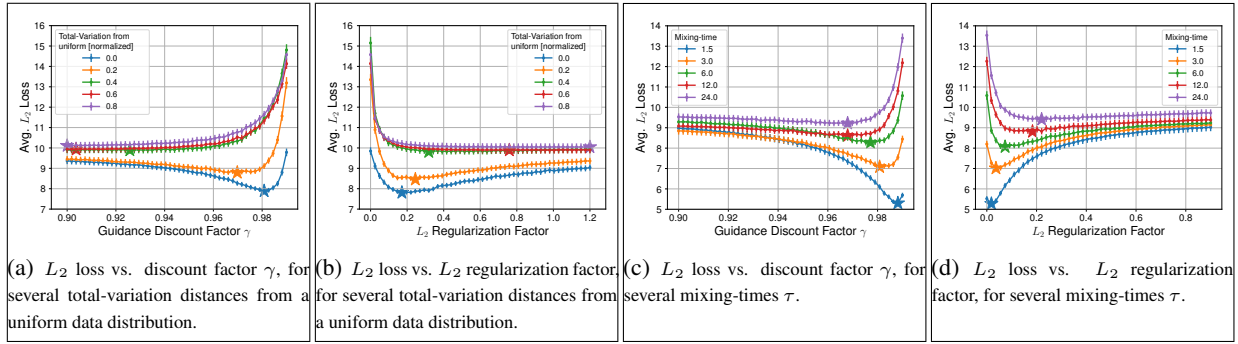

\centering
 \frame{
\subfigure[\scriptsize  {$L_2$ loss vs. discount factor $\gamma$, for several total-variation distances from a uniform data distribution.}]{
  \scalebox{\tabularPlotScale}{\input{images/2020_06_16_19_04_10_PolEval_LSTD_L2Loss_UniformityDist_DiscountReg.pgf}}
  \label{fig:PolEval_LSTD_L2Loss_UniformityDist_DiscountReg}
}}
 \frame{
\subfigure[\scriptsize  {$L_2$ loss vs. $L_2$ regularization factor, for several total-variation distances from a uniform data distribution.}]{
  \scalebox{\tabularPlotScale}{\input{images/2020_06_16_15_31_31__PolEval_LSTD_L2Loss_UniformityDist_L2Reg.pgf}}
  \label{fig:PolEval_LSTD_L2Loss_UniformityDist_L2Reg}
}}
 \frame{
\subfigure[\scriptsize  {$L_2$ loss vs. discount factor $\gamma$, for several mixing-times $\tau$.\newline}]{
  \scalebox{\tabularPlotScale}{\input{images/2020_06_17_17_16_21_PolEval_LSTD_L2Loss_MixTimes_DiscountReg.pgf}}
  \label{fig:PolEval_LSTD_L2Loss_MixTimes_DiscountReg}
}}
 \frame{
\subfigure[\scriptsize  {$L_2$ loss vs. $L_2$ regularization factor, for several mixing-times $\tau$.\newline}]{
  \scalebox{\tabularPlotScale}{\input{images/2020_06_17_16_58_17_PolEval_LSTD_L2Loss_MixTimes_L2Reg.pgf}}
  \label{fig:PolEval_LSTD_L2Loss_MixTimes_L2Reg}
}}
\caption{\textbf{Tabular Experiments - effect of data properties.} Loss vs. regularization factor for different regularizers, averaged over $1000$ MDP instances.
All results are with the LSTD algorithm.
The star shapes mark the minimum of the curve. Error bars represent $95\%$ confidence interval.}
\label{fig:Tabular_More_Exps}
\end{figure*}

 Figures \ref{fig:PolEval_LSTD_L2Loss_UniformityDist_DiscountReg} and \ref{fig:PolEval_LSTD_L2Loss_UniformityDist_L2Reg} show the loss when using each of the regularization methods, for various distances from a uniform distribution, when using the LSTD algorithm.
 As seen in the figure, for data distributions close to uniform, the benefit of regularization is greater.
In section \ref{sect:equiv} we predicted that discount regularisation will be more helpful for more uniform distributions. 
Interestingly, we find that the effectiveness of $L_2$ regularization is influenced in the same manner as discount regularization.

\paragraph{Mixing-time influence.}
Another interesting question regards the effect of the mixing-time on regularization effectiveness.
In Markov chains, the mixing-time describes the typical convergence time of the state distribution to the stationary distribution. 
It can be computed using the inverse spectral gap of the transition probabilities matrix \citep{levin2017markov, jerison2013general}.

To create trajectories with a specific mixing time we augmented the transition probabilities matrix to have the appropriate spectral gap for the specified mixing time (see full details in Appendix \ref{sect:mix_time_augment_procedure}).
We study a batch policy evaluation setting, where the behavioral policy is uniform, and the data is collected from two trajectories of length 50. The value is estimated using LSTD.
In each experiment repetition, we randomly create an MDP, derive the Markov process induced by a uniform policy, and apply the mixing time augmenting procedure.
 As seen in Figures \ref{fig:PolEval_LSTD_L2Loss_MixTimes_DiscountReg} and \ref{fig:PolEval_LSTD_L2Loss_MixTimes_L2Reg}, discount regularization and $L_2$ regularization are more effective in the slow mixing regime. 
 Intuitively, in this regime, limited data is less representative of the whole state space, which leads to higher estimation variance and so more regularization is needed.

\paragraph{Policy optimization.}
Improving performance of policy evaluation with regularization can improve performance of policy-iteration based algorithms. 
To demonstrate this, we run 5 episodes of approximate policy iteration: \textit{(i)} gather data by generating trajectories with $10$ time-steps by rolling out $\varepsilon$-greedy policy with $\varepsilon=0.1$, \textit{(ii)} run policy evaluation with SARSA, and \textit{(iii)} derive greedy policy w.r.t estimated value function.
The evaluation metric, \textit{optimality loss}, is the $L_1$ distance of the value of the learned policy $V^\pi$ to the value of the optimal policy  $V^*$,  $\norm{V^\pi - V^*}_1$, where the values are computed with the true model and $\gammaEval=0.99$.

\begin{figure}[t]
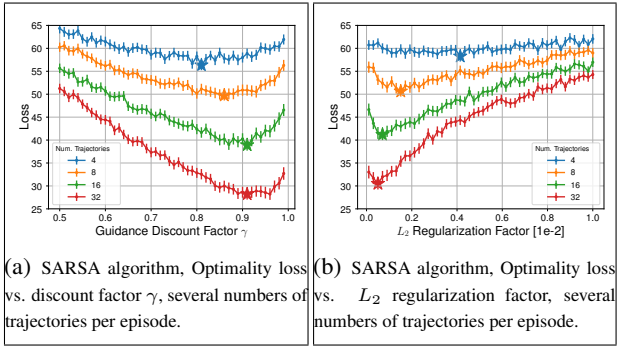

\centering
 \frame{
\subfigure[\scriptsize  {SARSA algorithm, Optimality loss vs. discount factor $\gamma$, several numbers of trajectories per episode.}]{
  \scalebox{\tabularPlotScale}{\input{images/2020_06_19_12_18_29_PolOpt_5_Episodes_DisountReg.pgf}}
  \label{fig:PolOpt_5_Episodes_DisountReg}
}}
 \frame{
\subfigure[\scriptsize  {SARSA algorithm, Optimality loss vs. $L_2$ regularization factor, several numbers of trajectories per episode.}]{
  \scalebox{\tabularPlotScale}{\input{images/2020_06_27_23_08_58__PolOpt_5_Episodes_L2Reg.pgf}}
  \label{fig:PolOpt_5_Episodes_L2Reg}
}}
\caption{\textbf{Tabular experiments - policy optimization.} Optimality loss vs. regularization factor for different regularizers, averaged over $1000$ MDP instances.
 The number of trajectories per episode is 16.
 The star shapes mark the minimum of the curve. Error bars represent $95\%$ confidence interval.}
\label{fig:Tabular_L2_Loss}
\end{figure}

\begin{figure}[t]
    \centering
    \includegraphics[width=\linewidth]{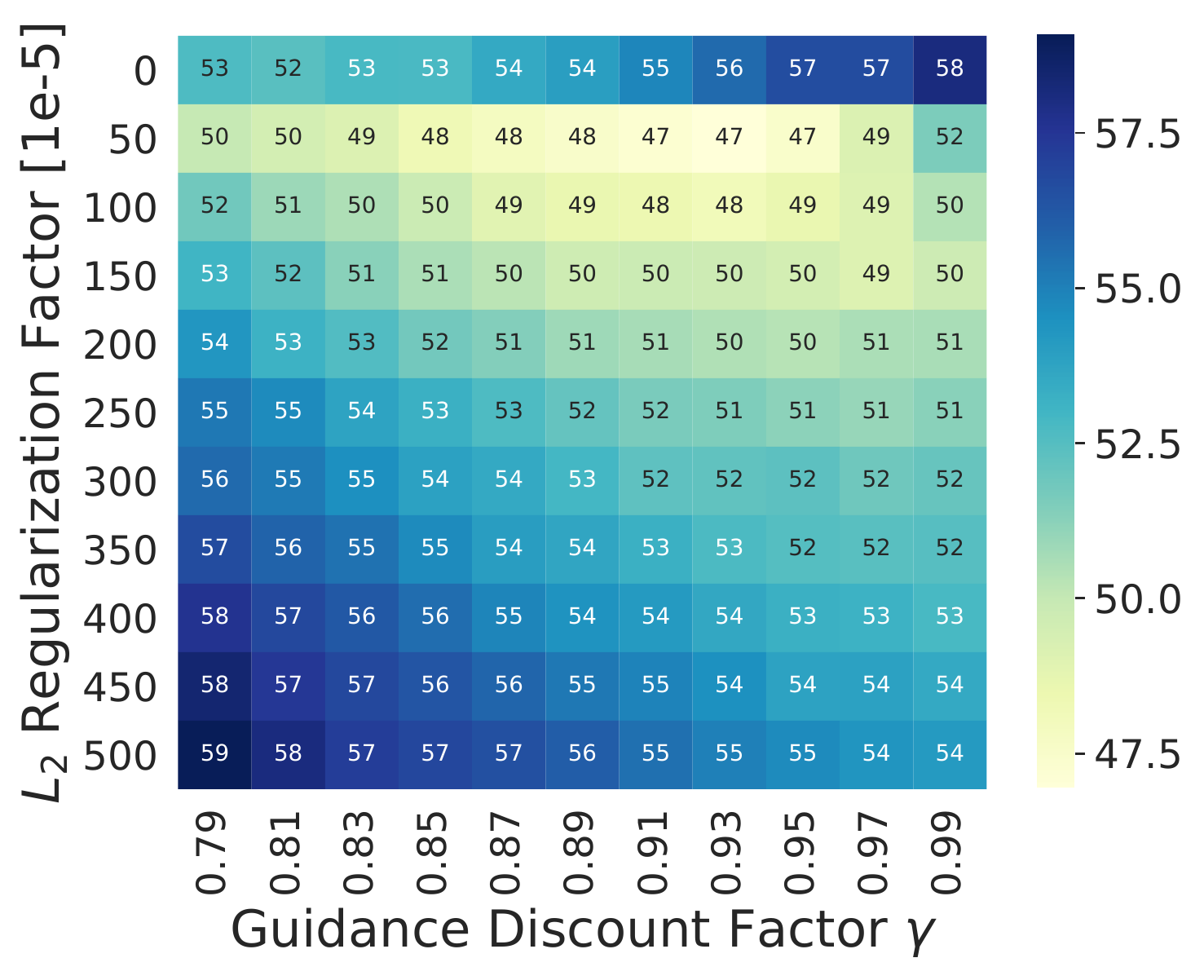}
  \caption{Optimality loss per guidance discount factor $\gamma$ and $L_2$ factor.
  Results for 5 episodes of policy-iteration with 8 trajectories of length 10 per episode.
The results are averaged across 1000 MDP instances from the environment.
The 95\% confidence interval is less than 2.5\% relative to the mean.}
\label{fig:2d_grid}
\end{figure}

In Figures \ref{fig:PolOpt_5_Episodes_DisountReg} and \ref{fig:PolOpt_5_Episodes_L2Reg} we see the results for discount and $L_2$ regularization respectively.
Both methods can achieve similar performance improvement.  As in previous experiments, when less data is available, stronger regularization is needed.
Note that while this experiment only tested one regularizer at a time, using a combination of both $L_2$ and discount regularization can considerably improve generalization, as seen in Figure \ref{fig:2d_grid}. 


\subsection{Deep RL Experiments} \label{sect:deep_experiments}

In this section, we investigate whether a reduced discount (or equivalently activation regularization) will benefit generalization from a finite sample in a continuous control with function approximation setting. 
Our experiments use the Mujoco environment \citep{todorov2012mujoco}.
To test the ability to generalize from finite data, we limited the number of time-steps from the environment to 200,000 or less. 


As a learning algorithm, we used the Twin Delayed DDPG (TD3) algorithm \citep{fujimoto2018addressing}, a recent actor-critic algorithm that achieves state-of-the-art performance in continuous control tasks. The policy evaluation stage of TD3 uses a variant of expected SARSA called target policy smoothing to estimate state-action values.
Similar experiments with the DDPG algorithm \citep{lillicrap2015continuous} are in Appendix \ref{sect:DDPG_results}.

All hyper-parameters are identical to those suggested by \citep{fujimoto2018addressing} except the following changes.
We tested with several amounts of total time-steps to simulate a limited data setting. 
As in \citet{fujimoto2018addressing}, The first $10^4$  time steps are used only for exploration. Another change to improve learning stability is increasing the batch size from $100$ to $256$. See Appendix \ref{sect:ImplementDetails}  for the complete implementation details.
We tested two regularization methods: \textit{(i)} \textit{discount regularization} -  $\gamma$ is varied and the $L_2$ factor is zero.
\textit{(ii)} \textit{$L_2$ regularization} - the $L_2$ factor  is varied and  $\gamma$  is fixed to high value of $0.999$.

Since the focus of this paper is regularization of the value estimation phase, we tested $L_2$ regularization only for the critic network.
As in common practice in deep learning, only the non-bias weight parameters are regularized and  since they are less prone to over-fitting \citep{goodfellow2016deep}.

For each tested hyper-parameter we repeated the experiments for $20$ different initial random seeds.  The averaging over a number of seeds allows for statistically significant results despite the high variance of the simulation environment \citep{henderson2018deep}.
In each repetition, the performance evaluation of the final policy is done by averaging the total undiscounted return (i.e, $\gammaEval = 1$) on $1000$ new episodes.

\newcommand{\mujocoScale}{0.25}  

\begin{figure*}[t!]
\subfigure[\scriptsize HalfCheetah-v2, discount regularization, 2e5 steps]{
  \scalebox{\mujocoScale}{\input{images/2020_05_14_22_34_20_HalfCheetah-v2_Discount_2e5.pgf}}
  \label{fig:td3_HalfCheetah_Discount_2e5}
}
\subfigure[\scriptsize{HalfCheetah-v2, $L_2$ regularization, 2e5 steps}]{
  \centering
    \scalebox{\mujocoScale}{\input{images/2020_05_24_19_06_32_HalfCheetah-v2_L2_2e5.pgf}}
  \label{fig:td3_HalfCheetah_L2_2e5}
  }
\subfigure[\scriptsize HalfCheetah-v2, discount regularization, 25e3 steps]{
\centering
  \scalebox{\mujocoScale}{\input{images/2020_05_14_22_34_20_HalfCheetah-v2_Discount_25e3.pgf}}
  \label{fig:td3_RandomMDP_L2_zero}
}
\subfigure[\scriptsize HalfCheetah-v2, $L_2$ regularization, 25e3 steps]{
\centering
  \scalebox{\mujocoScale}{\input{images/2020_05_24_19_06_32_HalfCheetah-v2_L2_25e3.pgf}}
  \label{fig:td3_HalfCheetah_L2_25e3}
 }
 \subfigure[\scriptsize Ant-v2, discount regularization, 2e5 steps]{
 \centering
  \scalebox{\mujocoScale}{\input{images/2020_05_09_22_01_18_Ant-v2_Discount_2e5.pgf}}
  \label{fig:td3_Ant_Discount_2e5}
}
\subfigure[\scriptsize{Ant-v2, $L_2$ regularization, 2e5 steps}]{
  \centering
    \scalebox{\mujocoScale}{\input{images/2020_05_24_19_17_52_Ant-v2_L2_2e5.pgf}}
  \label{fig:td3_Ant_L2_2e5}
  }
\subfigure[\scriptsize Ant-v2, discount regularization, 1e5 steps]{
\centering
  \scalebox{\mujocoScale}{\input{images/2020_05_09_22_01_18_Ant-v2_Discount_1e5.pgf}}
  \label{fig:td3_Ant_Discount_1e5}
}
\subfigure[\scriptsize Ant-v2, $L_2$ regularization, 1e5 steps]{
\centering
  \scalebox{\mujocoScale}{\input{images/2020_05_24_19_17_52_Ant-v2_L2_1e5.pgf}}
  \label{fig:td3_Ant_L2_1e5}
 }
   \subfigure[\scriptsize Hopper-v2, discount regularization, 2e5 steps]{
   \centering
  \scalebox{\mujocoScale}{\input{images/2020_05_09_21_59_07_Hopper-v2_Discount_2e5.pgf}}
  \label{fig:td3_Hopper_Discount_2e5}
}
\subfigure[\scriptsize{Hopper-v2, $L_2$ regularization, 2e5 steps}]{
  \centering
    \scalebox{\mujocoScale}{\input{images/2020_05_18_09_49_16_Hopper-v2_L2_2e5.pgf}}
  \label{fig:td3_Hopper_L2_2e5}
  } 
\subfigure[\scriptsize Hopper-v2, discount regularization, 5e4 steps]{
\centering
  \scalebox{\mujocoScale}{\input{images/2020_05_09_21_59_07_Hopper-v2_Discount_5e4.pgf}}
  \label{fig:td3_Hopper_Discount_5e4}
} 
\subfigure[ \scriptsize Hopper-v2, $L_2$ regularization, 5e4 steps]{
\centering
  \scalebox{\mujocoScale}{\input{images/2020_05_18_09_49_16_Hopper-v2_L2_5e4.pgf}}
  \label{fig:td3_Hopper_L2_5e4}
 }
  
 \caption{\textbf{Regularization in Mujoco experiments with limited data and TD3 algorithm.} Average total reward in evaluation episodes vs. regularization factor.  Results are averaged over $20$ simulations and $1000$ evaluation episodes. Shaded area represent $95\%$ confidence interval.}
 \label{fig:td3_Mujoco_reg}
\end{figure*}

The results appear in Figure \ref{fig:td3_Mujoco_reg}.
The results demonstrates that  discount regularization can lead to significant performance gain.
In the case of 200,000 time-steps, we can see that $\gamma$ values of around $0.99$ are optimal.
For lower numbers of time-steps, lower discount factors are generally more favourable. For example, in the Ant-v2 experiment $\gamma=0.8$ is optimal for 100,000 time-steps (Fig. \ref{fig:td3_Ant_Discount_1e5}).

If we compare $L_2$ regularization to  discount regularization, we see that sometimes it gives lower performance gain (e.g, Fig. \ref{fig:td3_Ant_Discount_2e5} and \ref{fig:td3_HalfCheetah_L2_2e5}), but in other cases it gives a higher gain, especially for smaller amount of time-steps (e.g., Fig. \ref{fig:td3_HalfCheetah_L2_25e3}  and \ref{fig:td3_Hopper_Discount_5e4}).

We note that there is a wide variability of behavior across the different Mujoco tasks (as has been observed also in previous work \citep{ahmed2019understanding}).
In practice, the discount factor should be chosen using a grid search for a specific environment and amount of available data.
However, our work suggests a few helpful guidelines: if less data is available, lower discounts become more favourable, in scenarios with non-uniform data coverage, or a fast mixing time, lowering the discount is likely to be less helpful.

Note that the common practice in actor-critic algorithms for learning Mujoco environments is to regularize the policy evaluation by setting $\gamma=0.99$ and $L_2$ factor of about $10^{-2}$ (e.g, \citet{lillicrap2015continuous}). Our results suggest that this hyper-parameter choice works well in some cases, but in other cases increasing the amount of regularization can significantly improve final performance.




\section{Related Work} \label{sect:Related}

     It is well-known that lower $\gamma$ increases convergence rate in many RL algorithms \citep{bertsekas1996neuro}, but several works showed that it can also improve final performance in the cae of limited data or approximation error.
	\citet{petrik2009biasing}  studied approximate dynamic-programming and showed that planning with a lower discount factor might be advised when the approximation error is large.
	\citet{chen2018improving} and \citet{franccois2019overfitting} studied similar phenomena in POMDPs.
	\citet{jiang2015dependence, jiang2016structural} studied a model-based RL setting and suggested that in the limited data regime, the performance of model-based RL can be improved by using a low discount factor in the planning phase.
    Our work identifies new elements that contribute to the effectiveness of discount regularization: uniformity and mixing rate. 
    
       In the planning setting, a classic result by \citet{blackwell1962discrete} shows that for every finite  MDP, there exists a discount factor $\gamma^*$ such that planing with any greater discount factor ($\gamma \geq \gamma^*$) leads to an optimal policy in the average reward sense.
   \cite{kakade2001optimizing} showed that for faster mixing MDPs, lowered discount factors introduces less bias int the average reward sense. Our work shows that in the learning setting, lowered discounts can even allow better generalization in faster mixing scenarios. 
   
   	The importance of regularization of generalization has also been demonstrated empirically with deep RL algorithms.  
	\citet{cobbe2018quantifying} suggested benchmarks for measuring generalization in deep RL and demonstrated that common regularization methods like $L_2$, can significantly improve generalization using the PPO algorithm \citep{schulman2017proximal}.
    \citet{farebrother2018generalization} showed regularization can improve generalization in Atari benchmarks when using the DQN algorithm \citep{mnih2015human}.
    \citet{parisi2019td} suggested a method for regularizing actor-critic algorithms by adding a TD error penalty in the actor’s objective.
    \citet{prokhorov1997adaptive} demonstrated the benefit of discount regularization using a schedule for increasing $\gamma$ as learning progresses. Similar scheduling is used in modern large scale RL applications \citep{OpenAI_dota}.
    \citet{xu2018meta} showed a gradient-based automatic hyper-parameter tuning method that achieved significant performance enhancement by tuning the discount.
    \citet{sherstan2019gamma} and \citet{romoff2019separating}  suggested methods for TD learning with a high discount via learning a sequence of value functions with lower discount factors.
    A recent line of works \citep{efroni2018beyond, tomar2019multi, tessler2020maximizing} proposes algorithmic schemes for using a small discount factor that asymptotically converge to the solution of the problem with the original discount. 

    While the benefits of a low discount factor have been shown in some settings, in other settings it has been shown to have adverse effects. The work of \citet{vanseijen2019using} analyze a family of small MDPs and show the existence of a sweet-spot in $\gamma$ selection.


\section{Conclusions}
In this paper, we studied the regularization effect of using a low discount factor in RL algorithms. In summary, our work
demonstrated empirically that discount regularization can significantly improve generalization performance when learning from limited data. 
In the tabular setting, we demonstrated that discount regularization is more effective for more uniform empirical state distribution or slower mixing rate.
In our experiments, discount and $L_2$ regularization had similar performance gain in the tabular settings, but different gains in the deep RL settings.

Our work opens several directions for further research.
\textit{(i)} Can theoretical results explain the phenomena  observed in our experiments?
\textit{(ii)} 
Can we explain the variation in performance between discount and $L_2$ regularization in the function approximation setting?
\textit{(iii)} 
Can we develop RL algorithms that utilize $L_2$ and discount regularization in an adaptive manner?



	\subsubsection*{Acknowledgments}

We thank Yonatan Efroni, Tom Zahavy, Nadav Merlis, Chen Tessler, Nir Baram, Ester Dorfman, Asaf Cassel, Guy Tennenholtz, Baruch Epstein, Tom Jurgenson, Alekh Agarwal, Tom Minka, Katja Hofmann and the Game Intelligence team at Microsoft Research, for helpful discussions of this work, and the anonymous reviewers for their helpful comments.  
The work of RM is partially supported by grant 451/17 from the Israel Science Foundation, by the Ollendorff Center of the Viterbi Faculty of Electrical Engineering at the Technion, and by the Skillman chair in biomedical sciences.

\bibliography{bib_file}
\bibliographystyle{icml2020}

\newpage 
\onecolumn
\appendix

\section{Appendix}

\subsection{Equivalence Proof for TD(0)} \label{sect:proof_td0}
In this section we present the proof of Proposition \ref{prop:Equiv_TD0}.

\begin{proof}
    Let $\phi$ be the sequence of parameters produced by Algorithm \ref{alg:TD0} when using discount factor $\gammaEval$, and added regularization function $\RegTerm(s, \theta) :=  \lambda \br*{\VhatT(s)}^2$, $\lambda := \frac{\gammaEval - \gamma}{2\gamma}$ , 
reward scaling $\xi:= \frac{\gammaEval}{\gamma}$,
learning rate  $\alpha_i' :=  \frac{\gamma}{\gammaEval} \alpha_i$
and same initial parameters $\theta_0$.
We will use induction to show $\theta_i = \phi_i, i=1,2,\dots$.

    The base case $\phi_0 =\theta_0$ follows immediately from the initialization.
    Assume $\phi_{i} = \theta_{i}$. 
    We now prove for $i + 1$.
    
    We can  rewrite $i$-th  step of Algorithm \ref{alg:TD0}  as
    \begin{align*}
       \theta_{i+1} & = \theta_i +  \alpha_i \left( r + \gamma \VhatTi(s') - \VhatTi(s) \right) \GradVhatTi(s) \nonumber \\  
        & \overset{(1)}{=}   \phi_i +  \alpha_i \left( r + \gamma  \VhatPi(s') -   \VhatPi(s) \right) \GradVhatPi(s) \nonumber \\  
         & =  \phi_i +  \alpha_i \frac{\gamma}{\gammaEval} \big( \frac{\gammaEval}{\gamma}  r + \gammaEval  \VhatPi(s') 
      - \frac{\gammaEval}{\gamma}  \VhatPi(s) \big)   \GradVhatPi(s) \nonumber \\    
      &  \overset{(2)}{=}   \phi_i +  \alpha'_i  \big( \frac{\gammaEval}{\gamma} r + \gammaEval\VhatPi(s') -   \VhatPi(s) +  \VhatPi(s) 
    - \frac{\gammaEval}{\gamma}  \VhatPi(s) \big) \GradVhatPi(s) \nonumber \\ 
        & =  \phi_i +  \alpha'_i  \left( \frac{\gammaEval}{\gamma} r + \gammaEval\VhatPi(s') -   \VhatPi(s) \right)   \GradVhatPi(s) 
        - \alpha_i' \frac{\gammaEval - \gamma}{\gamma}  \VhatPi(s) \GradVhatPi(s) \nonumber \\ 
        & = \phi_i  +  \alpha'_i  \left( \frac{\gammaEval}{\gamma} r + \gammaEval\VhatPi(s') -   \VhatPi(s) \right) \GradVhatPi(s) 
       - \alpha_i' \nabla \left( \frac{\gammaEval - \gamma}{2\gamma}   \br*{\VhatPi(s)}^2 \right) \nonumber \\ 
        & =\phi_{i+1} \nonumber \
    \end{align*}
      where equality (1) is due to the induction assumption and in (2) we defined $\alpha_i' :=  \alpha_i \frac{\gamma}{\gammaEval}$.
\end{proof}

\subsection{Equivalence for SARSA and Expected SARSA} \label{sect:Sarsa_Equiv}
In this section we prove an equivalence for the Expected SARSA(0) algorithm (Algorithm \ref{alg:SARSA0}), similarly to the proof for TD(0).
Same arguments apply for the vanilla SARSA algorithm (where ${\Vhat}_i(s')$ is replaced by $\QhatTi(s',a')$).

\begin{algorithm}
  \caption{Batch Expected SARSA(0)}
  \label{alg:SARSA0}
\begin{algorithmic}
  \STATE {\bfseries Hyper-parameters:}  $\gamma \in [0, \gammaEval]$  
  \STATE {\bfseries Input:}   $D$, $\pi$
    \FOR{ $i=0,1,..., \Niter - 1$}
       \STATE  Get uniformly random $(s,a,r,s')$ from $D$ 
       \STATE ${\Vhat}_i(s') := \sum_{a' \in \Actions} \pi(a'|s') \QhatTi(s',a') $
       \STATE  $
            \theta_{i+1} :=  \theta_{i}  +  \alpha_i [ r +  \gamma {\Vhat}_i(s')  - \QhatTi(s,a) ] \GradQhatTi(s,a)
            $
            \label{algStep:SARSA}
    \ENDFOR
\end{algorithmic} 
\end{algorithm}

\begin{algorithm}
  \caption{Batch Expected SARSA(0) with activation regularization}
  \label{alg:SARSA0_equiv}
\begin{algorithmic}
  \STATE {\bfseries Hyper-parameters:}   $\lambda \in \R^+$ (regularization factor),  $\xi \in \R^+$  (global reward scaling) 
  \STATE {\bfseries Input:}   $D$, $\pi$
    \FOR{ $i=0,1,..., \Niter - 1$}
       \STATE  Get uniformly random $(s,a,r,s')$ from $D$ 
       \STATE ${\Vhat}_i(s') := \sum_{a' \in \Actions} \pi(a'|s') \QhatPi(s',a') $
       \STATE  $
            \phi_{i+1} :=  \phi_{i}  +  \alpha_i' [\xi r +  \gammaEval {\Vhat}_i(s')  - \QhatPi(s,a) ] \GradQhatPi(s,a) - \alpha_i'  \nabla \br*{ \lambda \br*{\QhatPi(s,a)}^2}.
            $
            \label{algStep:SARSA_equiv}
    \ENDFOR
\end{algorithmic} 
\end{algorithm}

\begin{proposition} \label{prop:Equiv_SARSA}
Let $\theta_1, \theta_2,..$ be  the parameters produced by Algorithm \ref{alg:SARSA0}.
If  Algorithm \ref{alg:SARSA0_equiv} is run with initial parameters  $\phi_0 := \theta_0$ , step-size  $\alpha_i' :=  \frac{\gamma}{\gammaEval} \alpha_i$, reward scaling $\xi:= \frac{\gammaEval}{\gamma}$ and regularization factor  $\lambda := \frac{\gammaEval - \gamma}{2\gamma}$ then it produces the same sequence of parameters, i.e,  
  $\phi_k = \theta_k, k = 0,1,...$.
\end{proposition}

\begin{proof}
    We prove by induction.
    The base case $\phi_0 =\theta_0$ follows immediately from the initialization.
  
    Induction step: Assume $\phi_{i} = \theta_{i}$. 
    We now prove for $i + 1$.
    
    Using the induction assumption we can  rewrite $i$-th  step of Alg. \ref{alg:SARSA0}  as
    \begin{align*}
       \theta_{i+1} & = \theta_i +  \alpha_i \br*{r + \gamma \VhatTi(s') - \QhatTi(s,a)} \GradQhatTi(s,a) \nonumber \\  
        & =  \phi_i +  \alpha_i \br*{r + \gamma  \VhatPi(s') -   \QhatPi(s,a) }  \GradQhatPi(s,a) \nonumber \\    
          & =  \phi_i +  \alpha_i \frac{\gamma}{\gammaEval}  \br*{\frac{\gammaEval}{\gamma}  r + \gammaEval  \VhatPi(s') -  \frac{\gammaEval}{\gamma}  \QhatPi(s,a) }  \GradQhatPi(s,a) \nonumber \\    
      &  \overset{(1)}{=}  \phi_i +  \alpha'_i  \br*{\frac{\gammaEval}{\gamma} r + \gammaEval\VhatPi(s') -   \QhatPi(s,a) +  \QhatPi(s,a) -  \frac{\gammaEval}{\gamma}  \QhatPi(s,a) }    \GradQhatPi(s,a) \nonumber \\ 
        & =  \phi_i +  \alpha'_i  \br*{\frac{\gammaEval}{\gamma} r + \gammaEval\VhatPi(s') -  \QhatPi(s,a) }   \GradQhatPi(s,a)  - \alpha_i' \frac{\gammaEval - \gamma}{\gamma}  \QhatPi(s,a) \GradQhatPi(s,a) \nonumber \\ 
        & = \phi_i  +  \alpha'_i  \br*{\frac{\gammaEval}{\gamma} r + \gammaEval\VhatPi(s') -   \QhatPi(s,a) } \GradQhatPi(s,a)  - \alpha_i' \nabla \br*{ \frac{\gammaEval - \gamma}{2\gamma}   \br*{\QhatPi(s,a)}^2} \nonumber \\ 
        & =\phi_{i+1}, \nonumber \
    \end{align*}
   where in (1) we defined $\alpha_i' :=  \alpha_i \frac{\gamma}{\gammaEval}$.
\end{proof}

\subsection{The Equivalence $m$-step TD Prediction} \label{sect:EquivMstep}
In this section we will introduce a version of the equivalence for $m$-step TD updates.

The $m$-step TD update is defined as 
\begin{equation} \label{eq:basic_update_n_step}
    \theta \leftarrow \theta + \alpha_i \br*{\sum_{\tau=0}^{m-1} \gamma^\tau  r_\tau + \gamma^m \VhatT(s_n) - \VhatT(s)} \GradVhatT(s).
\end{equation}.

\begin{proposition} \label{prop:Equiv_n_step}
The semi-gradient $m$-step TD update step (\ref{eq:basic_update_n_step}) is equivalent to the following update step
    \begin{equation}  \label{eq:equiv_update_n_step}
       \theta \leftarrow \theta +  \alpha_i' \br*{   \sum_{\tau=0}^{m-1}  \gamma^\tau  \xi  r_\tau + \gammaEval^m \VhatT(s_m) - \VhatT(s) } \GradVhatT(s) - \alpha_i'  \nabla \br*{ \lambda \br*{\VhatT(s)}^2},
    \end{equation} 
    where $\alpha_i' :=  \alpha_i \frac{\gamma^m}{\gammaEval^m}$  is a modified step size, $\xi = \frac{\gammaEval^m}{\gamma^m}$ is a global reward scaling, and $ \lambda = \frac{\gammaEval^m - \gamma^m}{2\gamma^m}$ is a regularization factor.
    
\end{proposition}

\begin{proof}
    We can  rewrite the update of (\ref{eq:basic_update_n_step}) as:
    \begin{align*}
      &  \theta + \alpha_i \br*{\sum_{\tau=0}^{m-1} \gamma^\tau r_\tau + \gamma^m \VhatT(s_m) - \VhatT(s)} \GradVhatT(s)  \nonumber \\
      &  = \theta +  \alpha_i \frac{\gamma^m}{\gammaEval^m} \br*{\sum_{\tau=0}^{m-1} \gamma^\tau  \frac{\gammaEval^m}{\gamma^m} r_\tau + \gammaEval^m \VhatT(s_m) -  \frac{\gammaEval^m}{\gamma^m} \VhatT(s)} \GradVhatT(s)  \nonumber \\
             &  =\alpha_i \frac{\gamma^m}{\gammaEval^m} \br*{\sum_{\tau=0}^{m-1} \gamma^\tau  \frac{\gammaEval^m}{\gamma^m} r_\tau + \gammaEval^m \VhatT(s_m) -   \VhatT(s) + \br*{1-\frac{\gammaEval^m}{\gamma^m}} \VhatT(s)} \GradVhatT(s)  \nonumber 
    \end{align*}
    Denoting $\alpha_i' :=  \alpha_i \frac{\gamma^m}{\gammaEval^m}$  we can write
    \begin{align*}
                &  =  \alpha_i' \br*{\sum_{\tau=0}^{m-1} \gamma^\tau  \frac{\gammaEval^m}{\gamma^m} r_\tau + \gammaEval^m \VhatT(s_m) -   \VhatT(s)} + \alpha_i' \br*{1-\frac{\gammaEval^m}{\gamma^m}} \VhatT(s) \GradVhatT(s)  \nonumber \\
        & =  \alpha_i' \br*{\sum_{\tau=0}^{m-1} \gamma^\tau  \frac{\gammaEval^m}{\gamma^m} r_\tau + \gammaEval^m \VhatT(s_m) -   \VhatT(s)} - \alpha_i' \nabla  \br*{\frac{\gammaEval^m - \gamma^m}{2\gamma^m} \br*{\VhatT(s)}^2 }  \nonumber
    \end{align*}

\end{proof}
\subsection{The Equivalence for the LSTD Algorithm} \label{sect:EquivLSTD}
In this section we will introduce a version of the equivalence for the LSTD algorithm in the linear case.

Assume linear representation $\Vhat_\theta (s):= \phi(s)^\top \theta$. 
The input is a set of  transitions $\brc*{(s_i, a_i, r_i, s'_{i})}_{i=1}^N$
We use the $L_2$ regularized LSTD algorithm with a guidance discount factor $\gamma \leq \gammaEval$
The algorithm output is $\theta := A^{-1} b$ , where
$A := \frac{1}{N} \sum_{i=1}^{N} \phi(s_i) \br*{\phi(s_i) - \gamma \phi(s'_i)}^\top + \lambda I$, 
and $b := \frac{1}{N} \sum_{i=1}^{N}  r_i \phi(s_i)$.

We can re-write $A$ as follows
\begin{align*}
    A &=  \frac{1}{N} \sum_{i=1}^{N} \phi(s_i) \br*{\phi(s_i) - \gamma \phi(s'_i)}^\top  + \lambda I \\
    &= \frac{1}{N} \sum_{i=1}^{N} \phi(s_i) \br*{\phi(s_i) - \gammaEval \phi(s'_i)}^\top  
     + \br*{\gammaEval  - \gamma}  \frac{1}{N} \sum_{i=1}^{N} \phi(s_i)  \phi^\top(s'_i)  + \lambda I
\end{align*}

This shows that using a small discount $\gamma < \gammaEval$ in LSTD is equivalent to using a high discount $\gammaEval$ and adding an \textit{activation regularization}, $ \br*{\gammaEval  - \gamma}  \frac{1}{N} \sum_{i=1}^{N} \phi(s_i)  \phi^\top(s'_i) $.

As we saw for TD(0), in the case of orthonormal features (as in the tabular case with uniform visitation), we have an exact equivalence to an $L_2$ regularization term.

The exact same derivation can be done for the LSTDQ algorithm \citep{lagoudakis2003least}.

\subsection{Tabular Experiments - Additional Details}  \label{sect:additional_details_tabular}



\subsubsection{GridWorld environment details.} 
We constructed a $4 \times 4$ GridWorld environment.
For each instance of the MDP, 
a randomly chosen `goal' state is assigned to a high reward mean of $1$, while in all other
 states the reward mean is drawn uniformly from $[-0.5,0.5]$.
We assign the same reward mean for all actions at a given state.
The instantaneous reward signal is drawn from Gaussian with standard deviation $0.1$ and the state's reward mean. 
 The available actions at each state are \{`left', 'right', `up', `down', `stay'\}. 
 If the move is not valid, then the agent remains in the same state.
 Otherwise, the agent moves to the new sate with probability $p_s$, or otherwise stays in the current state. The probabilities $p_s,\forall s \in \States$ are drawn uniformly from $[0,1]$ when the MDP is created.
 In this problem there is no terminal state.

\subsubsection{Evaluation method details.}
We study the performance of the policy learned after on episode of approximate policy iteration.
In the first stage of the episode, a batch of transitions is collected.
Second, we run a batch policy evaluation algorithm to estimate the $Q$-function of the data collecting policy.
Third, we derive the greedy policy w.r.t.~$Q$, denoted $\pi$.
The performance of $\pi$ is measured by a loss function which is the $L_1$ distance of the value of the learned policy $V^\pi$ to the value of the optimal policy  $V^*$,  $\norm{V^\pi - V^*}_1$, where the values are computed with the true model and $\gammaEval$. 
We repeated the experiment for different numbers of samples collected in each episode. The results were averaged over $1000$ repetitions. 

\subsubsection{TD(0) Expected-SARSA Algorithm Details.}
In both algorithms we use large number of TD-iterations ($5000$) on the data set, where in each iterations we randomly sample a transition from the data set.
We use a large number of iterations since we are interested in evaluating the final performance and not the convergence rate.
The value function (or Q-function in Expected SARSA) is initialized with zero values.
The learning rate is $\alpha_i:=500/(1000+i)$, where $i$ is the iteration index.

\paragraph{Procedure for Augmenting the Mixing Time of a Markov Process}  \label{sect:mix_time_augment_procedure}

We describe the procedure we used for augmenting the transition probabilities matrix of a  Markov Process to have a specific mixing time.

\begin{itemize}
    \item Define the ‘target’ spectral gap according to the desired mixing time.

    \item Calculate the eigendecomposition of the transition matrix $P$.
    \item Force the desired spectral gap:
    \begin{itemize}
        \item Re-scale the magnitude  $\lambda_2$  to be according to the spectral gap.
        \item For each other eigenvalue that now has a higher magnitude than $\lambda_2$, re-scale it to be $\abs{\lambda_2}$.
\end{itemize}
\end{itemize}

\subsection{Additional Tabular Experiments} \label{sect:additional_experiments_tabular}

\subsubsection{Ranking Loss Evaluation} \label{sect:ranking_loss}

\begin{figure*}[t]
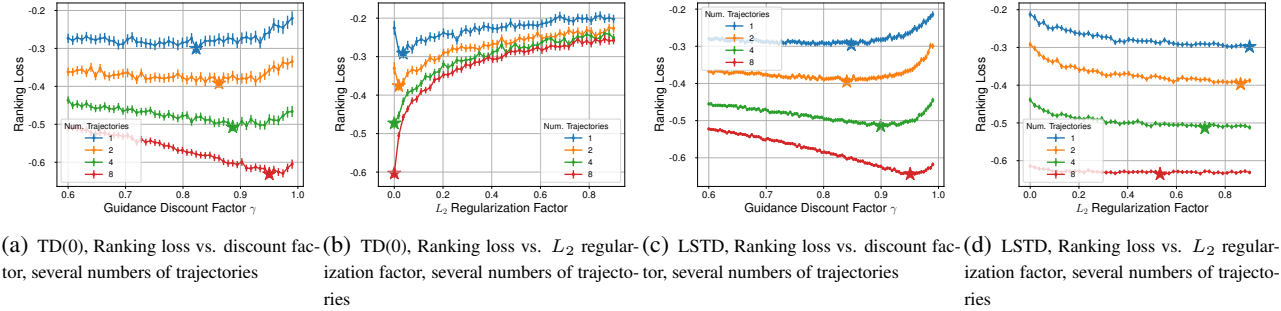

\centering
\subfigure[\scriptsize  {TD(0), Ranking loss vs. discount factor, several numbers of trajectories}]{
  \scalebox{\tabularPlotScale}{\input{images/2020_06_16_11_05_14_PolEval_TD_RankingLoss_DiscountReg.pgf}}
  \label{fig:PolEval_TD_RankingLoss_DiscountReg}
}
\subfigure[\scriptsize  {TD(0),  Ranking loss vs. $L_2$ regularization factor, several numbers of trajectories}]{
  \scalebox{\tabularPlotScale}{\input{images/2020_06_16_11_18_25_PolEval_TD_RankingLoss_L2Reg.pgf}}
  \label{fig:PolEval_TD_RankingLoss_L2Reg}
}
\subfigure[\scriptsize  {LSTD, Ranking loss vs. discount factor, several numbers of trajectories}]{
  \scalebox{\tabularPlotScale}{\input{images/2020_06_16_13_20_57_PolEval_LSTD_RankingLoss_DiscountReg.pgf}}
  \label{fig:PolEval_LSTD_RankingLoss_DiscountReg}
}
\subfigure[\scriptsize  {LSTD, Ranking loss vs. $L_2$ regularization factor, several numbers of trajectories}]{
  \scalebox{\tabularPlotScale}{\input{images/2020_06_16_12_24_45_PolEval_LSTD_RankingLoss_L2Reg.pgf}}
  \label{fig:PolEval_LSTD_RankingLoss_L2Reg}
}
\caption{\textbf{Tabular experiments with the ranking loss.} Loss vs. regularization factor for different regularizers, averaged over $1000$ MDP instances.
In each figure, the curves correspond to different number of samples per episode. The star shapes mark the minimum of the curve. Error bars represent $95\%$ confidence interval.}
\label{fig:Tabular_RankingLoss}
\end{figure*}

In Figure \ref{fig:Tabular_RankingLoss} we present the results with the ranking loss, corresponding to the results in Figure \ref{fig:Tabular_L2_Loss} of Section \ref{sect:tabular_emprical_demonstration} with the $L_2$ loss.
The ranking loss of the value estimation is defined by the negative Kendall's Tau correlation  \citep{kendall1948rank} between the rankings of the estimated and true value functions (evaluated with the evaluation discount factor $\gammaEval$). 

\subsection{Complete Implementation Details of Mujoco Experiments} \label{sect:ImplementDetails}
Our code uses the implementation of the TD3 and DDPG algorithms by \citet{fujimoto2018addressing}.
For completeness, we include here the full implementation details.

Critic Architecture
\begin{verbatim}
(state dim + action dim, 400)
ReLU
(action dim + 400, 300)
RelU
(300, 1)
\end{verbatim}

Actor Architecture
\begin{verbatim}
(state dim, 400)
ReLU
(400, 300)
RelU
(300, 1)
tanh
\end{verbatim}

\begin{table}
\centering
\caption{Hyper-parameters specification}
\begin{center}
\begin{small}
\begin{tabular}{lcc}
\toprule
\bf{Hyper-parameter} & \bf{Default Value} & \bf{Grid} \\
\midrule
Critic Learning Rate & $10^{-3}$ & - \\
Critic Regularization & None &\scriptsize{ $ \lambda_{L_2} \cdot ||\theta||^2$, $\lambda_{L_2} \in  \brc{0.,    0.005, 0.01, ....0.08 }$}  \\
Actor Learning Rate & $10^{-3}$ & - \\
Actor Regularization & None & - \\
Optimizer & Adam & - \\
Target Update Rate ($\tau$) & $5 \cdot 10^{-3}$ & - \\ 
Batch Size & $256$ &  -\\ 
Iterations per time step & $1$ & -\\
Discount Factor & $0.999$ & \scriptsize{$\gamma \in \{ 0.1, 0.2, ..., 0.9, 0.91, ..., 0.98, 0.985, 0.99, 0.995, 1. \}$  } \\ 
Reward Scaling & $1.0$ & - \\
Normalized Observations & False & - \\
Gradient Clipping & False & - \\
Exploration Policy & $\normal{0, 0.1}$ & - \\ 
\bottomrule
\end{tabular}
\end{small}
\end{center}
\end{table}

Each point is the parameter grid is averaged over $100$ random seeds. The final policy is evaluated by averaging $10$ episodes.
The computing infrastructure for running the experiments used 4 GeForce GTX 1080 GPUs.

\subsection{DDPG Algorithm Experiments} \label{sect:DDPG_results}
\begin{figure*}[t!]
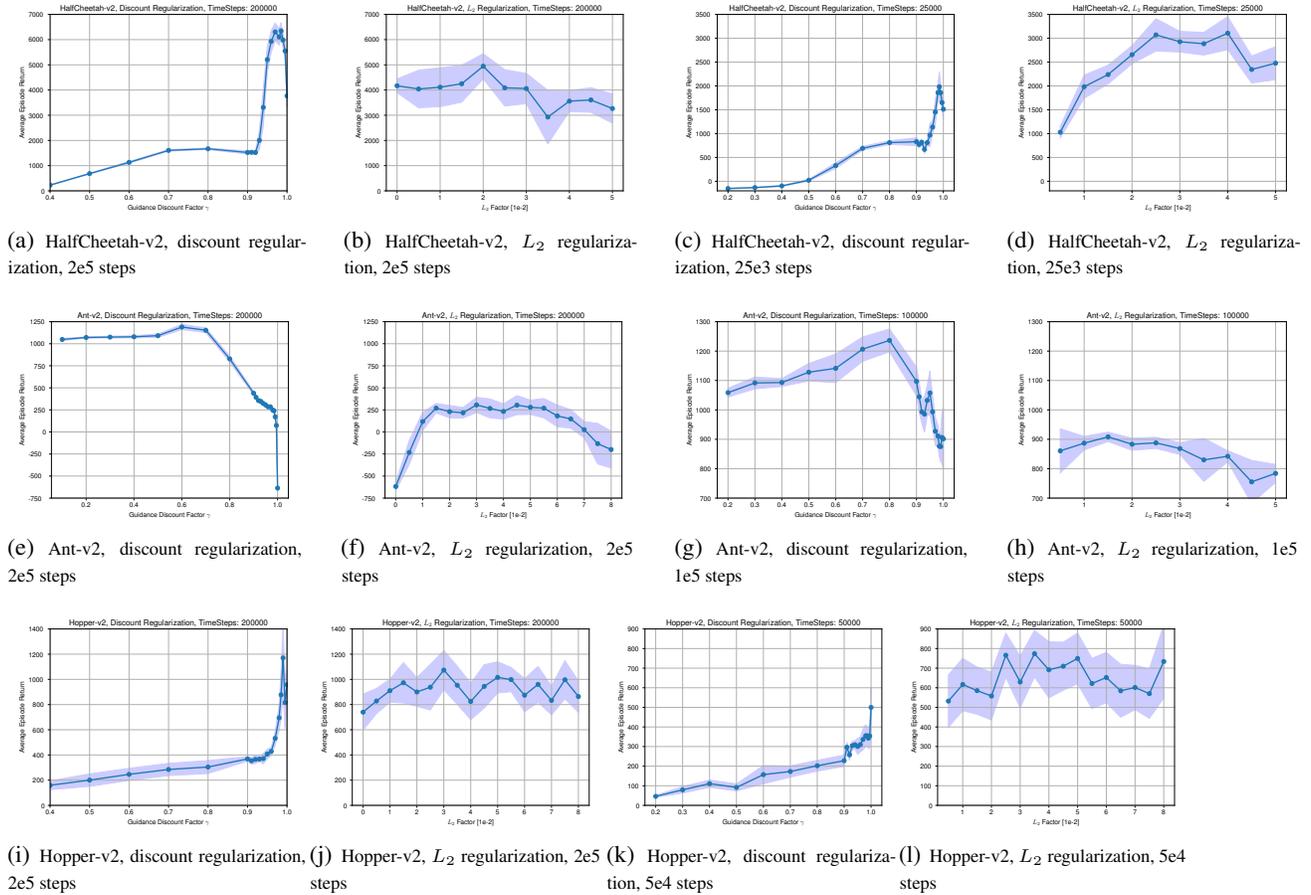

\subfigure[\scriptsize HalfCheetah-v2, discount regularization, 2e5 steps]{
  \scalebox{\mujocoScale}{\input{images/2020_06_12_23_46_47_HalfCheetah-v2_Discount_2e5_DDPG.pgf}}
  \label{fig:ddpg_HalfCheetah_Discount_2e5}
}
\subfigure[\scriptsize{HalfCheetah-v2, $L_2$ regularization, 2e5 steps}]{
  \centering
    \scalebox{\mujocoScale}{\input{images/2020_06_21_11_31_57_HalfCheetah-v2_L2_2e5_DDPG.pgf}}
  \label{fig:ddpg_HalfCheetah_L2_2e5}
  }
\subfigure[\scriptsize HalfCheetah-v2, discount regularization, 25e3 steps]{
  \scalebox{\mujocoScale}{\input{images/2020_06_12_23_46_47_HalfCheetah-v2_Discount_25e3_DDPG.pgf}}
  \label{fig:ddpg_RandomMDP_L2_zero}
}
\subfigure[\scriptsize HalfCheetah-v2, $L_2$ regularization, 25e3 steps]{
  \scalebox{\mujocoScale}{\input{images/2020_06_21_11_31_57_HalfCheetah-v2_L2_25e3_DDPG.pgf}}
  \label{fig:ddpg_HalfCheetah_L2_25e3}
 }
 \subfigure[\scriptsize Ant-v2, discount regularization, 2e5 steps]{
  \scalebox{\mujocoScale}{\input{images/2020_06_14_10_40_06_Ant-v2_Discount_2e5_DDPG.pgf}}
  \label{fig:ddpg_Ant_Discount_2e5}
}
\subfigure[\scriptsize{Ant-v2, $L_2$ regularization, 2e5 steps}]{
  \centering
    \scalebox{\mujocoScale}{\input{images/2020_06_18_14_24_49__Ant-v2_L2_2e5_DDPG.pgf}}
  \label{fig:ddpg_Ant_L2_2e5}
  }
\subfigure[\scriptsize Ant-v2, discount regularization, 1e5 steps]{
  \scalebox{\mujocoScale}{\input{images/2020_05_09_22_01_18_Ant-v2_Discount_1e5.pgf}}
  \label{fig:ddpg_Ant_Discount_1e5}
}
\subfigure[\scriptsize Ant-v2, $L_2$ regularization, 1e5 steps]{
  \scalebox{\mujocoScale}{\input{images/2020_05_24_19_17_52_Ant-v2_L2_1e5.pgf}}
  \label{fig:ddpg_Ant_L2_1e5}
 }
   \subfigure[\scriptsize Hopper-v2, discount regularization, 2e5 steps]{
  \scalebox{\mujocoScale}{\input{images/2020_06_14_17_32_16_Hopper-v2_Discount_2e5_DDPG.pgf}}
  \label{fig:ddpg_Hopper_Discount_2e5}
}
\subfigure[\scriptsize{Hopper-v2, $L_2$ regularization, 2e5 steps}]{
  \centering
    \scalebox{\mujocoScale}{\input{images/2020_06_21_00_14_27_Hopper-v2_L2_2e5_DDPG.pgf}}
  \label{fig:ddpg_Hopper_L2_2e5}
  }
\subfigure[\scriptsize Hopper-v2, discount regularization, 5e4 steps]{
  \scalebox{\mujocoScale}{\input{images/2020_06_14_17_32_16_Hopper-v2_Discount_5e4_DDPG.pgf}}
  \label{fig:ddpg_Hopper_Discount_5e4}
}
\subfigure[ \scriptsize Hopper-v2, $L_2$ regularization, 5e4 steps]{
  \scalebox{\mujocoScale}{\input{images/2020_06_21_00_14_27_Hopper-v2_L2_5e4_DDPG.pgf}}
  \label{fig:ddpg_Hopper_L2_5e4}
 }
  
 \caption{\textbf{Regularization in Mujoco experiments with limited data and DDPG algorithm.} Average total reward in evaluation episodes vs. regularization factor.  Results are averaged over $20$ simulations and $1000$ evaluation episodes. Shaded area represent $95\%$ confidence interval.}
 \label{fig:Mujoco_reg_ddpg}
\end{figure*}

In Figure \ref{fig:Mujoco_reg_ddpg} we present results for the DDPG algorithm, corresponding to the results described in Figure \ref{fig:td3_Mujoco_reg} of Section \ref{sect:deep_experiments} for the TD3 algorithm.

\end{document}